\documentclass{article}
\usepackage{comment}
\usepackage{spconf,amsmath,graphicx}
\usepackage{fancyhdr}
\usepackage{color}
\usepackage{array}
\usepackage{mdwmath}
\usepackage{diagbox}
\usepackage{mdwtab}
\usepackage{amsmath,amssymb}
\usepackage{cite}
\usepackage{graphicx}
\usepackage{glossaries}
\usepackage{listings}
\usepackage{subfig}
\usepackage{makecell}
\usepackage{ragged2e}
\usepackage[dvipsnames]{xcolor}
\usepackage{tabularx,ragged2e}
\usepackage[normalem]{ulem}
\newcolumntype{?}{!{\vrule width 2\arrayrulewidth}}
\usepackage{textcomp}{\tiny }
\usepackage{booktabs}
\usepackage{latexsym}
\usepackage{color}
\usepackage{url}
\usepackage{multirow}
\usepackage{bibentry}
\usepackage{enumitem}
\usepackage{lscape}
\usepackage{longtable}
\usepackage[utf8]{inputenc}
\usepackage{makecell}
\usepackage{mathrsfs}
\usepackage[mathscr]{euscript}
 \let\mathscr\relax
\usepackage[scr]{rsfso}
\usepackage{enumitem, kantlipsum}
\usepackage[dvipsnames]{xcolor}

\usepackage{flushend}

\def\Duno{\mathcal{D}_1}
\def\Ddue{\mathcal{D}_2}
\def\Dtre{\mathcal{D}_3}
\def\Dqua{\mathcal{D}_4}
\def\Dcin{\mathcal{D}_5}

\newacronym{cnn}{CNN}{Convolutional Neural Network}
\newacronym{gan}{GAN}{Generative Adversarial Network}
\newacronym{sg2ada}{StyleGAN2-ADA}{StyleGAN2 with Adaptive Discriminator Augmentation}
\newacronym{roc}{ROC}{Receiver Operating Characteristic}
\newacronym{auc}{AUC}{Area Under the Curve}
\newacronym{tpr}{TPR}{True Positive Rate}
\newacronym{fpr}{FPR}{False Positive Rate}

\title{Detecting GAN-generated Images by \\ Orthogonal Training of Multiple CNNs}

\name{Sara Mandelli, Nicol\`o Bonettini, Paolo Bestagini, Stefano Tubaro \vspace{-.8em}}
\address{Dipartimento di Elettronica, Informazione e Bioingegneria - Politecnico di Milano - Milan, Italy \\
\thanks{This material is based on research sponsored by the Defense Advanced Research Projects Agency (DARPA) and the Air Force Research Laboratory (AFRL) under agreement number FA8750-20-2-1004. The U.S. Government is authorized to reproduce and distribute reprints for Governmental purposes notwithstanding any copyright notation thereon. The views and conclusions contained herein are those of the authors and should not be interpreted as necessarily representing the official policies or endorsements, either expressed or implied, of DARPA and AFRL or the U.S. Government. This work was supported by the PREMIER project, funded by the Italian Ministry of Education, University, and Research within the PRIN 2017 program.}}
\begin{document}
\ninept
\sloppy
\maketitle
\begin{abstract}
In the last few years, we have witnessed the rise of a series of deep learning methods to generate synthetic images that look extremely realistic.
These techniques prove useful in the movie industry and for artistic purposes.
However, they also prove dangerous if used to spread fake news or to generate fake online accounts.
For this reason, detecting if an image is an actual photograph or has been synthetically generated is becoming an urgent necessity.
This paper proposes a detector of synthetic images based on an ensemble of Convolutional Neural Networks (CNNs).
We consider the problem of detecting images generated with techniques not available at training time.
This is a common scenario, given that new image generators are published more and more frequently.
To solve this issue, we leverage two main ideas: (i) CNNs should provide ``orthogonal'' results to better contribute to the ensemble; (ii) original images are better defined than synthetic ones, thus they should be better trusted at testing time.
Experiments show that pursuing these two ideas improves the detector accuracy on NVIDIA's newly generated StyleGAN3 images, never used in training.

\end{abstract}
\begin{keywords}
Image forensics, synthetic images, GAN, CNN
\end{keywords}

\vspace{-.25em}
\section{Introduction}
\label{sec:introduction}

Over the last few years, we assisted in an escalation of methods for the production of increasingly more realistic 
synthetically generated images \cite{karras2018, sg2, Karras2020ada, song2021scorebased}. 
The first architectures produced blurry and low-resolution images with a general lack of details.
Recently, giant steps have been made to raise the bar and overcome those issues.
This is evident by the recent release of a new \gls{gan} architecture by NVIDIA, namely StyleGAN3~\cite{Karras2021}, which produces high-quality images that can easily fool human eyes.

On the one hand, the authors of image generators put much effort into generating very realistic pictures.
On the other hand, they are aware of the variety of problems an overly realistic architecture can create.
Generated images can be used over social media for many malicious intents, from scams to identity stealing, and the general public is not ready to face this menace.
A recent work~\cite{lago2021} shows how is it difficult for humans to tell real and generated faces apart when they have just a few seconds to make the decision.
The interviewed people rated StyleGAN2~\cite{sg2} images as real in the $68\%$ of the cases, whereas real images were rated as real only in the $52\%$ of the cases.
Similarly, the study conducted in \cite{nightingale2022ai} shows how synthetic faces prove even more trustworthy at human inspection.
Notice that these studies do not even consider the more recent StyleGAN3 yet.

Given these premises, it is evident that being able to detect if an image is a natural photograph or it has been 
synthetically
generated is becoming a task of utmost importance.
For this reason, the multimedia forensics community is developing a series of techniques to solve the 
synthetically
generated image detection problem.
Some methods are based on hand-crafted features fed to classifiers \cite{marra2019gans, barni2020cnn, bonettini2021icpr, mandelli2021forensic}.
A wide variety of solutions prefer a purely data-driven approach based on training an end-to-end detector \cite{wang2020cnn, cozzolino2021towards}.
However, many of these techniques tend to suffer in classifying images that deviate from the characteristics of their training set.
Unfortunately, it is nowadays impractical to assume that the characteristic of any 
synthetically
generated image can be perfectly known at training stage, as new image generation techniques are developed continuously.
For this reason, the latest research trend is to develop methods that can generalize well, detecting images generated with unseen techniques \cite{gragnaniello2021gan, mandelli2021forensic}.

In this paper, we tackle the problem of 
\gls{gan}-generated image detection.
This is, given an image under analysis, to detect if it is a real photograph or it has been synthetically generated by a \gls{gan}.
We consider
the realistic and challenging scenario in which test images may come from generators that were unknown to the analyst at training time.
To solve the \gls{gan}-generated image detection problem, we propose an ensemble of \glspl{cnn}.

The proposed method leverages two main ideas to increase the robustness of unseen generated images.
First, \glspl{cnn} contributing to the ensemble should be as much ``orthogonal'' as possible.
We propose a training strategy that increases the diversity among the different learners for this purpose.
This prevents the \glspl{cnn} from overfitting the image generators used in training, thus enabling the ensemble to take a better decision on newly generated images.
Second, the detection problem is better defined over real images than synthetic ones.
Indeed, it is safer to assume that the analyst can train on a broad set of real photographs that better represent the real-image class.
On the contrary, it is hard to assume that the analyst can train on synthetic images generated with all the possible existing techniques, as these change and get updated too frequently in time.
Therefore, we propose a score aggregation strategy that better favour decisions towards the real-image class.

Our experimental campaign is designed to test the proposed training and aggregation strategies on top of a baseline \gls{cnn} detector based on EfficientNet \cite{tan2019efficientnet}.
We show that our technique is able to: (i) better draw the separation line between real and synthetic images by separating the score distributions of the two classes; (ii) accurately detect StyleGAN3 images as \gls{gan}-generated, even though they have never been used in training.
\vspace{-1em}

\section{Proposed method}
\label{sec:method}



We can summarize the main objectives of the \gls{gan}-generated image detection problem into three primary tasks:
(i) generalize very well to new \glspl{gan} unseen during training phase; (ii) be robust against post-processing operations applied on images; (iii) achieve a missed detection rate (i.e., the number of synthetic images detected as real) as low as possible.
To improve \gls{gan} generalization and robustness to editing operations, we propose a procedure based on orthogonal training of multiple \glspl{cnn}, all based on the same backbone architecture but each trained on a different training dataset. 
At the testing stage, we propose a patch selection and aggregation strategy that considerably reduces the missed detection rate. 

\begin{figure}[t]
\centering
\includegraphics[width=.95\columnwidth]{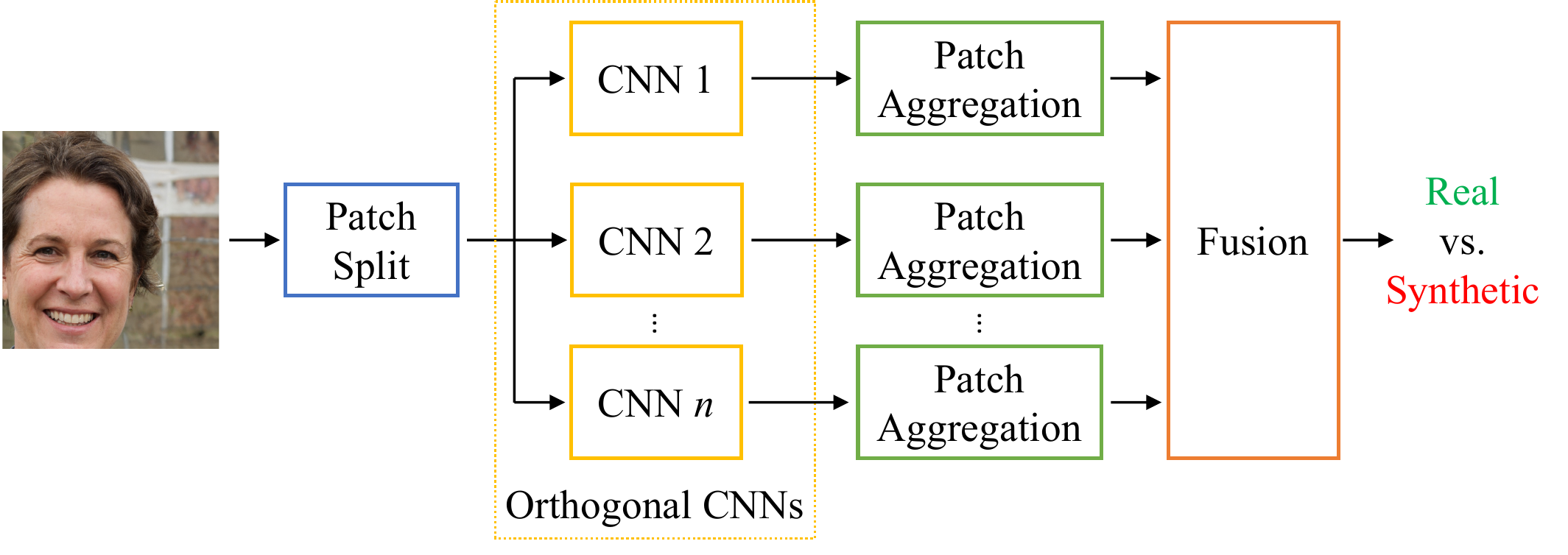}
\caption{\footnotesize{Proposed method for \gls{gan} image detection. Given an image, we split it into patches which are fed to multiple orthogonal \glspl{cnn}. We aggregate the patch scores to obtain a single image score per \gls{cnn}. Eventually, we fuse the scores into the final image score to classify the image as real or synthetic.}}
\label{fig:pipeline}
\end{figure}

Fig.~\ref{fig:pipeline} reports the sketch of the proposed testing pipeline. 
In a nutshell, given a query image, we classify it as being real or synthetically generated by selecting several patches from it and passing them through multiple orthogonal \glspl{cnn}. Then, we aggregate the patch scores and fuse the \glspl{cnn} results into a single prediction associated with the entire image.
We provide more details about the proposed approach in the following lines.
\vspace{-1em}

\subsection{Orthogonal Training}
\label{subsec:orth_training}

To improve the generalization of the proposed method against new \glspl{gan}, we train multiple \glspl{cnn} over different training datasets, which are ``orthogonal'' one another (with a slight abuse of terminology).
For clarity's sake, 
we consider two datasets ``orthogonal'' if one of the following conditions is met:
\begin{itemize}[leftmargin=*]
    \item the datasets include images depicting different semantic content (e.g., cats or humans);
    \item the datasets include images that underwent different post-processing (e.g., uncompressed or compressed images);
    \item the datasets include images that underwent different compressions (e.g., different JPEG implementations);
    \item the datasets include images synthesized by different \glspl{gan}.
\end{itemize}
The key idea of the proposed training orthogonalization is that every single \gls{cnn} should capture slightly different traces with respect to the others.
The ensemble of many \glspl{cnn} trained in an orthogonal fashion proves to achieve improvements with respect to training a unique \gls{cnn} over the whole entirety of data at disposal.

The common backbone used for all the proposed \glspl{cnn} is the EfficientNet-B4 model \cite{tan2019efficientnet}, well known in the computer vision and multimedia forensics communities due to the outstanding results achieved in many tasks although requiring few network parameters \cite{bonettini2021icpr}. 
Each \gls{cnn} works at patch level, always considering squared RGB patches of $N \times N$ pixels as input and providing a single score per patch.
Considering an ensemble of $C$ \glspl{cnn} analyzing $P$ patches per image, we define as $\hat{y}^c_p$ the score estimated by the $c$-th \gls{cnn} for the $p$-th patch, 
with $c \in [1, C]$ and $p \in [1, P]$.

To improve the robustness against various post-processing operations, we apply strong data augmentations as suggested in many state-of-the-art works for synthetic image detection \cite{gragnaniello2021gan, cozzolino2021towards}.
The list of possible augmentations emulates common editing operations that can be applied by amateur users when retouching their photographs. Moreover, malicious users could also apply the same operations to hide the traces left by the synthetic generation process. We consider horizontal and vertical flip, random $90$-degree rotation, histogram equalization, random blur, random changes in brightness, contrast, color and saturation, random downscale and upscale, and finally JPEG Compression with quality factors randomly selected from $30$ to $100$. Each augmentation is applied with probability $50\%$, except for JPEG compression, which is applied with probability $70\%$. The parameters are those defined in \cite{Buslaev2020}.

At testing stage, for each \gls{cnn}, we obtain different scores associated with the patches extracted from the query image. Real and synthetic patches are associated with negative and positive scores, respectively.
We fuse these scores by following the aggregation strategy presented in the next section to obtain the final image score.
\vspace{-1em}

\subsection{Patch Aggregation Strategy}
\label{subsec:patch_agg}


Given a test image, every orthogonal \gls{cnn} returns many scores associated with the patches extracted from the image.
When fusing the patch scores, we aim at reducing the detection errors on the synthetic images. Thus, the missed detection rate is the most critical parameter to maintain as low as possible.

The proposed approach is based on the consideration that, when training a generic \gls{gan} detector, it is reasonable to assume that the characteristics of real images are easier to be captured than those of synthetic ones.
Indeed, everybody could collect a set of original photographs and assign them the label ``real'', whereas collecting a sufficiently vast and various synthetic dataset might be more elaborate and certainly requires a little expertise.
Moreover, contrarily to the ``real'' class, the ``synthetic'' one is constantly and rapidly evolving, as new proposed methodologies for generating synthetic content emerge every day, not limited to \glspl{gan} only \cite{song2021scorebased, Dhariwal2021}.
Given these premises, we can reasonably assume that many \gls{cnn} detectors trained over orthogonal datasets might correctly classify a real query image with a high precision level, as long as they are accurately trained. We cannot make the same assumption for synthetic query images, as they might be generated from novel unseen \glspl{gan}.

Here comes the proposed patch aggregation strategy.
When a query image passes from a \gls{cnn} detector and \textit{all} its extracted patches are classified as real, the \gls{cnn} classifies the entire image as real.
If \textit{at least one} patch among those extracted from the test image is detected as synthetic, the \gls{cnn} assigns the entire image to the synthetic-image class.
In particular, the \gls{cnn} score associated with the image is the best score achieved among all the patches for what concerns the detected class.
Since real and synthetic images are associated with negative and positive scores, respectively, the image is assigned the minimum score among the patches if the detected class is real-image. Otherwise, we assign the maximum score.
Formally, the image score by the $c$-th \gls{cnn} is defined as
\begin{equation}
    \hat{y}^c = \begin{cases}
    \min_p{\hat{y}^c_p} & \text{if} \quad \hat{y}^c_p < 0, \; \forall p \in [1, P] \\
    \max_p{\hat{y}^c_p} & \text{otherwise}
    \end{cases}.
\end{equation}

Eventually, we equally weight the orthogonal \glspl{cnn} to assign the global image score, which is the arithmetic mean among all the image scores returned by the networks, i.e., 
$\hat{y} = \frac{1}{C} \sum_{c=1}^C \hat{y}^c$.
\vspace{-1em}
\section{Experimental analysis}
\label{sec:experiments}

\subsection{Dataset}
\label{subsec:dataset}

We perform our investigations over the dataset used for the competition recently organized by NVIDIA on StyleGAN3 Synthetic Image Detection \cite{nvidia_challenge} within the DARPA's SemaFor program.
The purpose was to simulate an open-world setting in which 
new unseen \glspl{gan} (e.g., StyleGAN3 \cite{Karras2021}) should be detected.

The real class of the testing data consisted of images selected from three public datasets: the FFHQ \cite{ffhq} (depicting human faces taken from photographs), the Metfaces \cite{metfaces} (depicting human faces taken from works of art), and the AFHQ2 \cite{afhq} (including photographs of animal faces from three domains of cat, dog, and wildlife). 
The synthetic images to be tested were all generated through the recently released StyleGAN3 network
\cite{Karras2021}, trained on real images selected from the three previously reported datasets.
Every real dataset corresponds to two possible synthetic versions of it, the version \textit{r} and the version \textit{t}, according to the specific StyleGAN3 configuration chosen at generation stage. 
The images from Metfaces and AFHQ2 datasets did not undergo post-processing or compression, while a few synthetic images from the FFHQ dataset underwent compression and resizing. 

The competition did not pose any limit on the kind of training data to be used for developing the proposed \gls{gan} detector, except for removing from the training data the real images belonging to the testing dataset and every synthetic image generated through StyleGAN3.
Given these premises, our testing dataset coincides with the testing dataset of the NVIDIA competition.
The training dataset consists of different datasets, purposely built so to implement the orthogonal \gls{cnn} training described in Section~\ref{subsec:orth_training}. 
Every orthogonal dataset is exploited for training an EfficientNet-B4, working with squared RGB patches of size $128 \times 128$. 
Following our previous considerations, we build $5$ ``orthogonal'' datasets: 

\noindent\textbf{Dataset $\Duno$.} This dataset includes all the real images from FFHQ, Metfaces and AFHQ2 available for training ($ \sim 116 K$). The synthetic images ($ \sim 200 K$) are selected from the synthetic versions of the three datasets, generated through state-of-the-art models for synthetic image generation (i.e., StyleGAN2 \cite{sg2}, StarGAN-v2 \cite{afhq}, Taming Transformers \cite{taming_tx}, FaceVid2Vid \cite{facev2v} and Score-based models \cite{song2021scorebased}).
During training, the images undergo multiple augmentations from the list reported in Section \ref{subsec:orth_training}, JPEG compression included. Then, $1$ patch per image is randomly selected and fed to the \gls{cnn}.

\noindent\textbf{Dataset $\Ddue$.}
This dataset includes the same real and synthetic images exploited for $\Duno$, with the difference that here we first randomly extract $1$ patch per image, then we apply the same augmentations defined for $\Duno$. From the point of view of the post-processing applied, $\Duno$ is orthogonal to $\Ddue$, especially for the JPEG compression. Indeed, by construction, all the patches from $\Ddue$ are aligned to the $8 \times 8$ pixel grid introduced by JPEG compression, while in $\Duno$ the patches can have any random alignment. As already shown in \cite{mandelli2020wifs}, taking care of the JPEG grid alignment is of paramount importance for multimedia forensics tasks. The datasets $\Duno$ and $\Ddue$ allow exploring this issue for the \gls{gan} detection problem.

\noindent\textbf{Dataset $\Dtre$.}    
This dataset includes only the real images from AFHQ2 available for training ($\sim14 K$) and an equal number of their synthetic versions generated through StyleGAN2 and StarGAN-v2. $10$ random patches are extracted per image, then undergo all the augmentations, except for JPEG compression. 
$\Dtre$ focuses only on one semantic category (i.e., the animal faces), on a few \glspl{gan} and is entirely orthogonal to $\Duno$ and $\Ddue$ for what regards JPEG compression.

\noindent\textbf{Dataset $\Dqua$.} 
Ideally, this dataset would include only the real images from Metfaces available for training ($\sim 2 K$) and an equal number of their synthetic versions generated through StyleGAN2. This would
guarantee complete semantic orthogonality with respect to $\Dtre$. Actually, the training process was unstable due to the very limited number of Metfaces images. To augment the dataset dimensions, we decided to include in 
$\Dqua$ also the AFHQ2-related images. We extract
$10$ random patches per image and apply augmentations, except for JPEG compression. As well as $\Dtre$, $\Dqua$ is entirely orthogonal to $\Duno$ and $\Ddue$ for what concerns JPEG compression. 

\noindent\textbf{Dataset $\Dcin$.} 
This dataset includes only real images from FFHQ available for training ($\sim 100 K$) and almost $170K$ synthetic versions of them generated through StyleGAN2, Taming Transformers, FaceVid2Vid and Score-based models. We randomly extract $1$ patch per image and pass it through the augmentations, JPEG compression included. $\Dcin$ is entirely orthogonal to $\Dtre$ and $\Dqua$ concerning the semantic content and is partially orthogonal for the \glspl{gan} used. Moreover, $\Dcin$ is entirely orthogonal to $\Duno$ about JPEG alignment.

At deployment stage, we extract RGB patches $128 \times 128$ from the query image in different ways according to the \gls{cnn} to be fed. For the \gls{cnn} trained over $\Duno$, we randomly select $200$ patches per image. For the remaining \glspl{cnn}, we always feed them with around $180$ patches per image, aligned with the $8 \times 8$ pixel grid introduced by JPEG compression.
This operation is done to ensure that the potential editing traces undergone by the test patches match with the JPEG training augmentations. Indeed, for building $\Duno$, the training patches can be misaligned to the JPEG grid, while the remaining datasets always match the JPEG grid alignment, if JPEG is present.
\vspace{-1em}

\subsection{Experimental setup}
\label{subsec:setup}


We keep $80\%$ of the training images for training phase, leaving the remaining $20\%$ for the validation.
As commonly done in \gls{cnn} training, we initialize the network weights using those trained on the ImageNet database.
Every \gls{cnn} is trained using cross-entropy loss and Adam optimizer with default parameters for a maximum of $500$ epochs.
The learning rate is initialized to $0.001$ and is decreased by a factor $10$ if the loss does not decrease for $10$ epochs.
Training is stopped if the loss does not improve for more than $20$ epochs, then the model providing the best validation loss is selected.
The experimental code is available at \url{https://github.com/polimi-ispl/GAN-image-detection}.
\vspace{-1em}

\subsection{Results}
\label{subsec:results}
This section reports the results achieved by the performed experimental campaign. 
First,  we show the performance of the proposed patch aggregation strategy, then we evaluate the orthogonal \gls{cnn} training. Eventually, we compare our results with state-of-the-art.

\noindent\textbf{Patch aggregation.} 
To show the effectiveness of the proposed patch aggregation strategy, we compare our approach with standard patch aggregation methodologies.
For brevity's sake, we show the benefits of our strategy only on the results achieved by one single \gls{cnn}, as the trend is the same for all the considered networks.
\begin{figure*}[t]
\centering
\includegraphics[width=.9\textwidth]{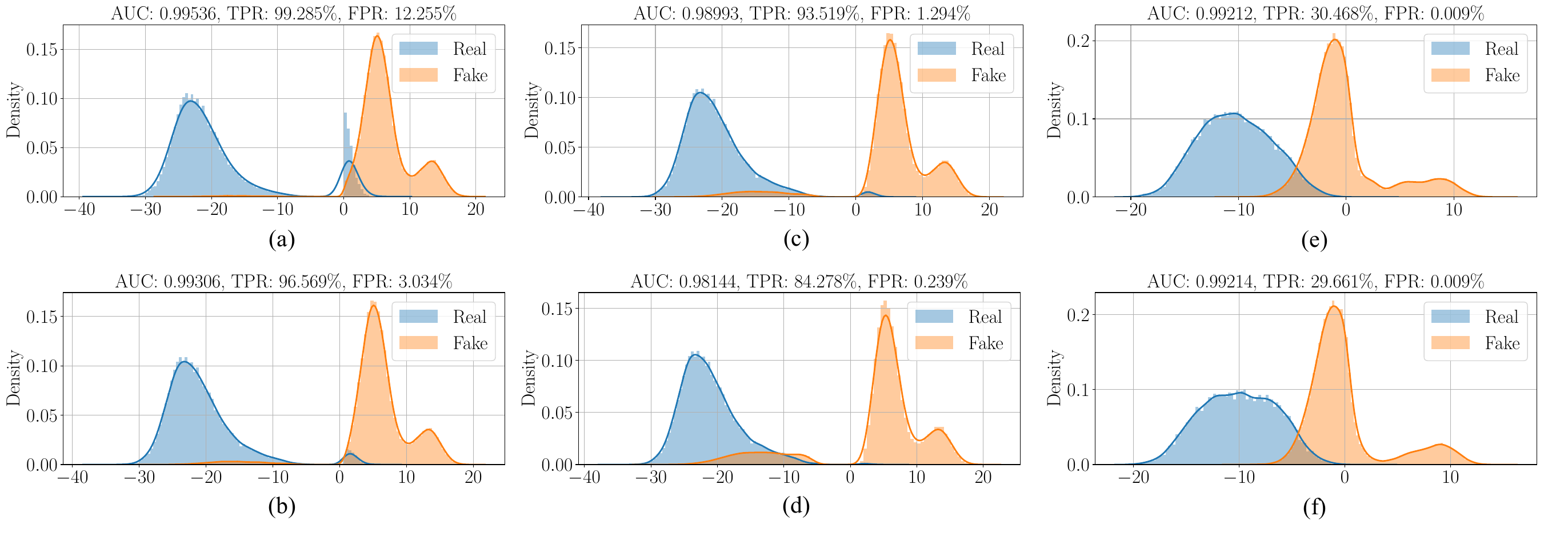}
\vspace{-1em}
\caption{\footnotesize{Histogram of the image scores achieved by the \gls{cnn} trained on dataset $\Ddue$: (a) reports the scores obtained by following the proposed patch aggregation strategy; (b), (c) and (d) show the scores obtained by modifying the threshold 
on the number of ``synthetic'' patches for assigning the image label 
to $5$, $10$ and $25$, respectively; (e) and (f) report the results if we aggregate the patch scores by computing their arithmetic mean and median, respectively.}}
\label{fig:patch_agg}
\end{figure*}
\begin{table*}[h!]
\caption{\footnotesize{AUC achieved in the eight considered testing scenarios and over the global test dataset. In bold, the two best AUCs for each scenario.}}
\label{tab:binary_results_auc}
\centering
\resizebox{0.9\textwidth}{!}{
 \begin{tabular}{@{}cccccccccc@{}}
 \toprule
   & AFHQ2-r & AFHQ2-t & Metfaces-r & Metfaces-t & FFHQ-r & FFHQ-t & FFHQ-r, res-comp & FFHQ-t, res-comp & Global
\\\midrule
$\textrm{CNN}_1$ & $\mathbf{0.9971}$ & $0.9995$ & $0.9777$ & $0.9896$ & $\mathbf{0.9999}$ & $0.9995$ & $0.9932$ & $0.9932$ & $0.9951$\\  
$\textrm{CNN}_2$ & $0.9884$ & $0.9954$ & $0.8996$ & $0.9347$ & $0.9999$ & $\mathbf{0.9997}$ & $\mathbf{0.9997}$ & $\mathbf{0.9997}$ & $\mathbf{0.9954}$\\  
$\textrm{CNN}_3$ & $0.9954$ & $\mathbf{0.9996}$ & $0.9909$ & $0.9834$ & $0.9999$ & $0.9994$ & $0.4218$ & $0.4313$ & $0.8092$\\  
$\textrm{CNN}_4$ & $0.9755$ & $0.9986$ & $\mathbf{0.9982}$ & $\mathbf{0.9991}$ & $0.9999$ & $0.9988$ & $0.8432$ & $0.8534$ & $0.9456$\\ 
$\textrm{CNN}_5$ & $0.6021$ & $0.5858$ & $0.6185$ & $0.7322$ & $0.9998$ & $0.9995$ & $\mathbf{0.9997}$ & $\mathbf{0.9997}$ & $0.9682$\\  \midrule
Fusion & $\mathbf{0.9991}$ & $\mathbf{0.9999}$ & $\mathbf{0.9919}$ & $\mathbf{0.9964}$ & $\mathbf{0.9999}$ & $\mathbf{0.9999}$ & ${0.9995}$ & ${0.9995}$ & $\mathbf{0.9995}$\\  \bottomrule
\end{tabular}
}
\vspace{-1em}
\end{table*}

Fig.~\ref{fig:patch_agg} depicts the achieved image scores' distributions by aggregating the patch scores returned by the \gls{cnn} trained on $\Ddue$. In particular, Fig.~\ref{fig:patch_agg}(a) reports the results of the proposed method: if at least one patch is detected as synthetic, the image is assigned the ``best'' score among the synthetic ones. Figs.~\ref{fig:patch_agg}(b)-(c)-(d) show the results obtained by modifying this strict condition, letting the number of patches required for assigning the label ``synthetic'' grow to $5$, $10$ and $25$ patches, respectively. 
Figs~\ref{fig:patch_agg}(e)-(f) report the results obtained by selecting the arithmetic mean and the median among the patch scores, respectively.
For each scenario, we report the corresponding \gls{auc} of the \gls{roc} curve and the \gls{tpr} and \gls{fpr} achieved in the confusion matrix.

Our approach achieves the highest \gls{auc} and an extremely high detection accuracy for synthetic images at the cost of a few false alarms.
As we increase the number of patches detected as synthetic for assigning the final score to the image (see Figs.~\ref{fig:patch_agg}(b)-(c)-(d)), the number of false alarms reduces, but the missed detections also increase. The arithmetic mean and median of the patch scores are far from being competitive with the proposed method.

\noindent\textbf{Orthogonal CNN training.} 
Table~\ref{tab:binary_results_auc} reports the results of every single \gls{cnn} and of the ensemble.
We show the \gls{auc} achieved on the global test set, but we also investigate different scenarios in which only the real images of a particular dataset (e.g., FFHQ, Metfaces or AFHQ2) are compared with their synthetic versions generated through StyleGAN3.
The considered scenarios are: (i) real AFHQ2 vs. synthetic AFHQ2 generated with \textit{r} configuration; (ii) real AFHQ2 vs. synthetic AFHQ2 generated with \textit{t} configuration; (iii) real Metfaces vs. synthetic Metfaces, \textit{r}-versions; (iv) real Metfaces vs. synthetic Metfaces, \textit{t}-versions; (v) real FFHQ vs. synthetic FFHQ, \textit{r}-versions, without resizing and compression; (vi) real FFHQ vs. synthetic FFHQ, \textit{t}-versions, without resizing and compression;
(vii) real FFHQ vs. real FFHQ, \textit{r}-versions, with resizing and compression; (viii) real FFHQ vs. synthetic FFHQ, \textit{r}-versions, with resizing and compression.

The \gls{cnn} ensemble often reports the best results. 
Regarding the single \glspl{cnn}, the best methods on average are those trained over $\Duno$ and $\Ddue$. This was expected, as these \glspl{cnn} were trained over a larger and more various amount of data with respect to the last three of them. However, every orthogonal training dataset carries important contributions related to the specific kind of data it is focused on. 
For instance, $\textrm{CNN}_3$ and $\textrm{CNN}_4$ achieve extremely high \glspl{auc} on AFHQ2 and Metfaces datasets, respectively. $\textrm{CNN}_5$ achieves almost perfect \glspl{auc} over FFHQ undergone post-processing operations.

All \glspl{cnn} report acceptable results in the global test scenario. Nonetheless, due to their specific training implementation, some \glspl{cnn} might be more prone to detection errors than others in particular test scenarios, whereas their ensemble always maintains robust. Aiming at simulating realistic situations in which test images come from unknown generative models, the ensemble of multiple \glspl{cnn} proves to be a valid option for synthetic image detection, paving the way towards robust and generalized solutions.

\noindent\textbf{Comparison with state-of-the-art.} 
The proposed \gls{gan} detector ranked first in the competition organized by NVIDIA, outperforming the results achieved by many expert teams in the field of multimedia forensics. 
Indeed, our method achieved the highest \gls{auc} over the global test set, as well as the best results in all the eight testing scenarios described previously.
We refer the interested reader to \cite{nvidia_challenge} for any additional details and for comparing the state-of-the-art results.
\vspace{-1em}


\section{Conclusions}
\label{sec:conclusions}

In this paper we proposed a synthetic image detector based on an ensemble of \glspl{cnn}, which
are trained to increase the diversity within the ensemble.
Our score aggregation strategy takes into account the fact that some image generators can be unknown at training time.
Results show that these ideas help improving the detector accuracy on StyleGAN3 images that have never been used for training.

Despite the promising results, the orthogonality among the trained \glspl{cnn} is only empirically verified at test time by observing the detector accuracy.
Future work will be devoted to a deeper study of the \glspl{cnn} diversity from a more theoretical view point.
This will enable the development of ad-hoc training strategies.

\bibliographystyle{IEEEbib}
\bibliography{biblio}

\begin{thebibliography}{10}

\bibitem{karras2018}
T.~Karras, T.~Aila, S.~Laine, and J.~Lehtinen,
\newblock ``Progressive growing of {GAN}s for improved quality, stability, and
  variation,''
\newblock in {\em International Conference on Learning Representations}, 2018.

\bibitem{sg2}
T.~Karras, S.~Laine, M.~Aittala, J.~Hellsten, J.~Lehtinen, and T.~Aila,
\newblock ``Analyzing and improving the image quality of {StyleGAN},''
\newblock in {\em IEEE/CVF Conference on Computer Vision and Pattern
  Recognition (CVPR)}, 2020.

\bibitem{Karras2020ada}
T.~Karras, M.~Aittala, J.~Hellsten, S.~Laine, J.~Lehtinen, and T.~Aila,
\newblock ``{Training Generative Adversarial Networks with Limited Data},''
\newblock in {\em Conference on Neural Information Processing Systems
  (NeurIPS)}, 2020.

\bibitem{song2021scorebased}
Y.~Song, J.~Sohl-Dickstein, D.~P. Kingma, A.~Kumar, S.~Ermon, and B.~Poole,
\newblock ``Score-based generative modeling through stochastic differential
  equations,''
\newblock in {\em International Conference on Learning Representations}, 2021.

\bibitem{Karras2021}
T.~Karras, M.~Aittala, S.~Laine, E.~H\"ark\"onen, J.~Hellsten, J.~Lehtinen, and
  T.~Aila,
\newblock ``Alias-free generative adversarial networks,''
\newblock in {\em Conference on Neural Information Processing Systems
  (NeurIPS)}, 2021.

\bibitem{lago2021}
F.~Lago, C.~Pasquini, R.~Böhme, H.~Dumont, V.~Goffaux, and G.~Boato,
\newblock ``More real than real: A study on human visual perception of
  synthetic faces,''
\newblock {\em IEEE Signal Processing Magazine}, vol. 39, pp. 109--116, 2022.

\bibitem{nightingale2022ai}
S.~J. Nightingale and H.~Farid,
\newblock ``{AI}-synthesized faces are indistinguishable from real faces and
  more trustworthy,''
\newblock {\em Proceedings of the National Academy of Sciences (PNAS)}, vol.
  119, 2022.

\bibitem{marra2019gans}
F.~Marra, D.~Gragnaniello, L.~Verdoliva, and G.~Poggi,
\newblock ``Do gans leave artificial fingerprints?,''
\newblock in {\em 2019 IEEE Conference on Multimedia Information Processing and
  Retrieval (MIPR)}, 2019.

\bibitem{barni2020cnn}
M.~Barni, K.~Kallas, E.~Nowroozi, and B.~Tondi,
\newblock ``Cnn detection of gan-generated face images based on cross-band
  co-occurrences analysis,''
\newblock in {\em 2020 IEEE International Workshop on Information Forensics and
  Security (WIFS)}, 2020.

\bibitem{bonettini2021icpr}
N.~Bonettini, E.~D. Cannas, S.~Mandelli, L.~Bondi, P.~Bestagini, and S.~Tubaro,
\newblock ``Video face manipulation detection through ensemble of {CNNs},''
\newblock in {\em International Conference on Pattern Recognition (ICPR)},
  2021.

\bibitem{mandelli2021forensic}
S.~Mandelli, D.~Cozzolino, E.~D. Cannas, J.~P. Cardenuto, D.~Moreira,
  P.~Bestagini, W.~Scheirer, A.~Rocha, L.~Verdoliva, S.~Tubaro, and E.~J. Delp,
\newblock ``Forensic analysis of synthetically generated scientific images,''
\newblock {\em arXiv preprint arXiv:2112.08739}, 2021.

\bibitem{wang2020cnn}
S.-Y. Wang, O.~Wang, R.~Zhang, A.~Owens, and A.~A. Efros,
\newblock ``Cnn-generated images are surprisingly easy to spot... for now,''
\newblock in {\em IEEE/CVF Conference on Computer Vision and Pattern
  Recognition (CVPR)}, 2020.

\bibitem{cozzolino2021towards}
D.~Cozzolino, D.~Gragnaniello, G.~Poggi, and L.~Verdoliva,
\newblock ``Towards universal gan image detection,''
\newblock in {\em International Conference on Visual Communications and Image
  Processing (VCIP)}, 2021.

\bibitem{gragnaniello2021gan}
D.~Gragnaniello, D.~Cozzolino, F.~Marra, G.~Poggi, and L.~Verdoliva,
\newblock ``Are gan generated images easy to detect? a critical analysis of the
  state-of-the-art,''
\newblock in {\em IEEE International Conference on Multimedia and Expo (ICME)},
  2021.

\bibitem{tan2019efficientnet}
M.~Tan and Q.~Le,
\newblock ``Efficientnet: Rethinking model scaling for convolutional neural
  networks,''
\newblock in {\em International Conference on Machine Learning (ICML)}, 2019.

\bibitem{Buslaev2020}
A.~Buslaev, V.~I. Iglovikov, E.~Khvedchenya, A.~Parinov, M.~Druzhinin, and
  A.~A. Kalinin,
\newblock ``Albumentations: fast and flexible image augmentations,''
\newblock {\em Information}, vol. 11, pp. 125, 2020.

\bibitem{Dhariwal2021}
P.~Dhariwal and A.~Q. Nichol,
\newblock ``{Diffusion Models Beat GANs on Image Synthesis},''
\newblock in {\em Conference on Neural Information Processing Systems
  (NeurIPS)}, 2021.

\bibitem{nvidia_challenge}
``{NVIDIA} {StyleGAN3} synthetic image detection,''
  \url{https://github.com/NVlabs/stylegan3-detector}, 2021.

\bibitem{ffhq}
``{Flickr-Faces-HQ} dataset ({FFHQ}),''
  \url{https://github.com/NVlabs/ffhq-dataset}.

\bibitem{metfaces}
``Metfaces dataset,'' \url{https://github.com/NVlabs/metfaces-dataset}.

\bibitem{afhq}
Y.~Choi, Y.~Uh, J.~Yoo, and J.-W. Ha,
\newblock ``{StarGAN}v2: Diverse image synthesis for multiple domains,''
\newblock in {\em IEEE/CVF Conference on Computer Vision and Pattern
  Recognition (CVPR)}, 2020.

\bibitem{taming_tx}
P.~Esser, R.~Rombach, and B.~Ommer,
\newblock ``Taming transformers for high-resolution image synthesis,''
\newblock in {\em Proceedings of the IEEE/CVF Conference on Computer Vision and
  Pattern Recognition (CVPR)}, 2021.

\bibitem{facev2v}
T.-C. Wang, A.~Mallya, and M.-Y. Liu,
\newblock ``One-shot free-view neural talking-head synthesis for video
  conferencing,''
\newblock in {\em IEEE/CVF Conference on Computer Vision and Pattern
  Recognition (CVPR)}, 2021.

\bibitem{mandelli2020wifs}
S.~Mandelli, N.~Bonettini, P.~Bestagini, and S.~Tubaro,
\newblock ``Training cnns in presence of jpeg compression: Multimedia forensics
  vs computer vision,''
\newblock in {\em IEEE International Workshop on Information Forensics and
  Security (WIFS)}, 2020.

\end{thebibliography}

\end{document}